\pdfoutput=1

\documentclass[11pt]{article}

\usepackage[]{acl}
\usepackage{authblk}

\usepackage{times}
\usepackage{latexsym}

\usepackage{multirow}

\usepackage[T1]{fontenc}

\usepackage[utf8]{inputenc}

\usepackage{microtype}

\usepackage{inconsolata}

\usepackage{graphicx}
\usepackage{booktabs}
\usepackage{amsmath}
\usepackage{xspace}
\usepackage{subfigure}
\title{Exploring the Role of Transliteration in In-Context Learning for Low-resource Languages Written in Non-Latin Scripts}

\author[]{\bf Chunlan Ma$^{\text *}$}
\author[]{\bf Yihong Liu$^{\text *}$}
\author[]{\bf Haotian Ye$^{\text *}$}
\author[]{\bf Hinrich Sch\"utze}

\affil{Center for Information and Language Processing, LMU Munich \\ Munich Center for Machine Learning (MCML)
 \protect\\ \texttt{\{chunlan, yihong,  yehao\}@cis.lmu.de}}

\begin{document}
\maketitle

\def\combined{\textsc{Script}$_{\{\text{Combined}\}}$\xspace}
\def\original{\textsc{Script}$_{\{\text{Orig}\}}$\xspace}
\def\transliteration{\textsc{Script}$_{\{\text{Latn}\}}$\xspace}

\def\secref#1{\S\ref{sec:#1}}
\def\seclabel#1{\label{sec:#1}}

\newcounter{notecounter}
\newcommand{\enotesoff}{\long\gdef\enote##1##2{}}
\newcommand{\enoteson}{\long\gdef\enote##1##2{{
\stepcounter{notecounter}
{\large\bf
\hspace{1cm}\arabic{notecounter} $<<<$ ##1: ##2
$>>>$\hspace{1cm}}}}}
\enoteson

\begin{abstract}

Decoder-only large language models (LLMs) excel in high-resource languages across various tasks through few-shot or even zero-shot in-context learning (ICL).
However, their performance often does not transfer well to low-resource languages, especially those written in non-Latin scripts.
Inspired by recent work that leverages transliteration in encoder-only models, we investigate whether transliteration\footnote{We consider a special type of transliteration that converts non-Latin scripts into Latin script (also called romanization).} is also effective in improving LLMs' performance for low-resource languages written in non-Latin scripts.
To this end, we propose three prompt templates, where the target-language text is represented in (1) its original script (\original), (2) Latin script (\transliteration), or (3) both (\combined).
We apply these methods to several representative LLMs of different sizes on various tasks including text classification and sequential labeling.
Our findings show that the effectiveness of transliteration varies by task type and model size. For instance, all models benefit from transliterations for sequential labeling (with increases of up to 25\%).
We make our code publicly available.



\end{abstract}

\section{Introduction}


Decoder-only LLMs, such as LLaMA \citep{touvron2023llama}, Mixtral \citep{jiang2024mixtral}, XGLM \citep{lin2022fewshot}, and BLOOM \citep{workshop2023bloom}, have shown impressive capability across a wide range of tasks for high-resource languages, particularly through few-shot ICL \citep{Brown2020fewshot}.
However, they often underperform in low-resource languages, especially those written in underrepresented scripts.
Multiple reasons exist, such as the scarcity of low-resource languages in the training data \citep{nllbteam2022language,üstün2024aya}, insufficient crosslingual alignment during pretraining \citep{hämmerl2024understanding}, as well as English being the only language in the instruction tuning phase \citep{chen-etal-2024-monolingual}.
The mainstream methodology attempts to address this issue by translating the texts written in languages other than English into English using either external machine translation systems \citep{artetxe-etal-2023-revisiting} or self-translate, i.e., translation by leveraging the few-shot translation capabilities of the model itself \citep{etxaniz2023multilingual}.
However, the quality of translations is constrained by the quality of the external systems or the LLM itself.
Additionally, this type of approach is infeasible for truly low-resource languages.

\begin{figure}
\setlength{\belowcaptionskip}{-0.5cm}
\centering
\subfigure{\includegraphics[width=0.49\textwidth]{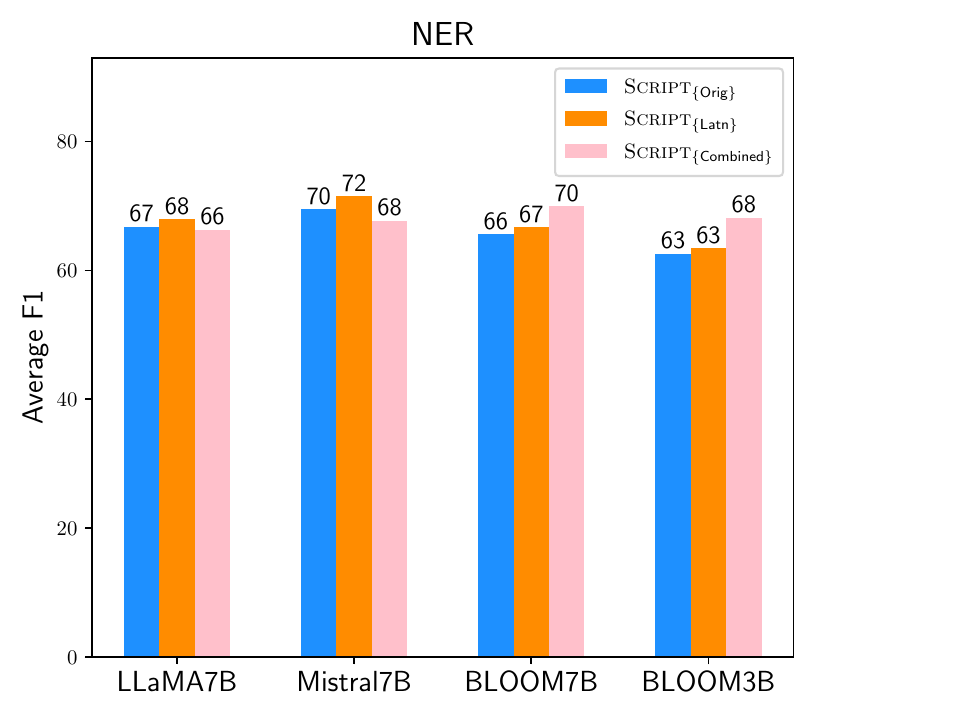}}
\hspace{-0.2cm}
\caption{Results of LLaMA7B, Mistral7B, BLOOM7B and BLOOM3B on NER task. By leveraging transliteration, \transliteration or \combined consistently improve the performance on NER across all models.
}
\label{fig:first_page_figure}
\end{figure}


\begin{figure*}
    \setlength{\belowcaptionskip}{-0.3cm}
  \centering
  \includegraphics[width=\textwidth]{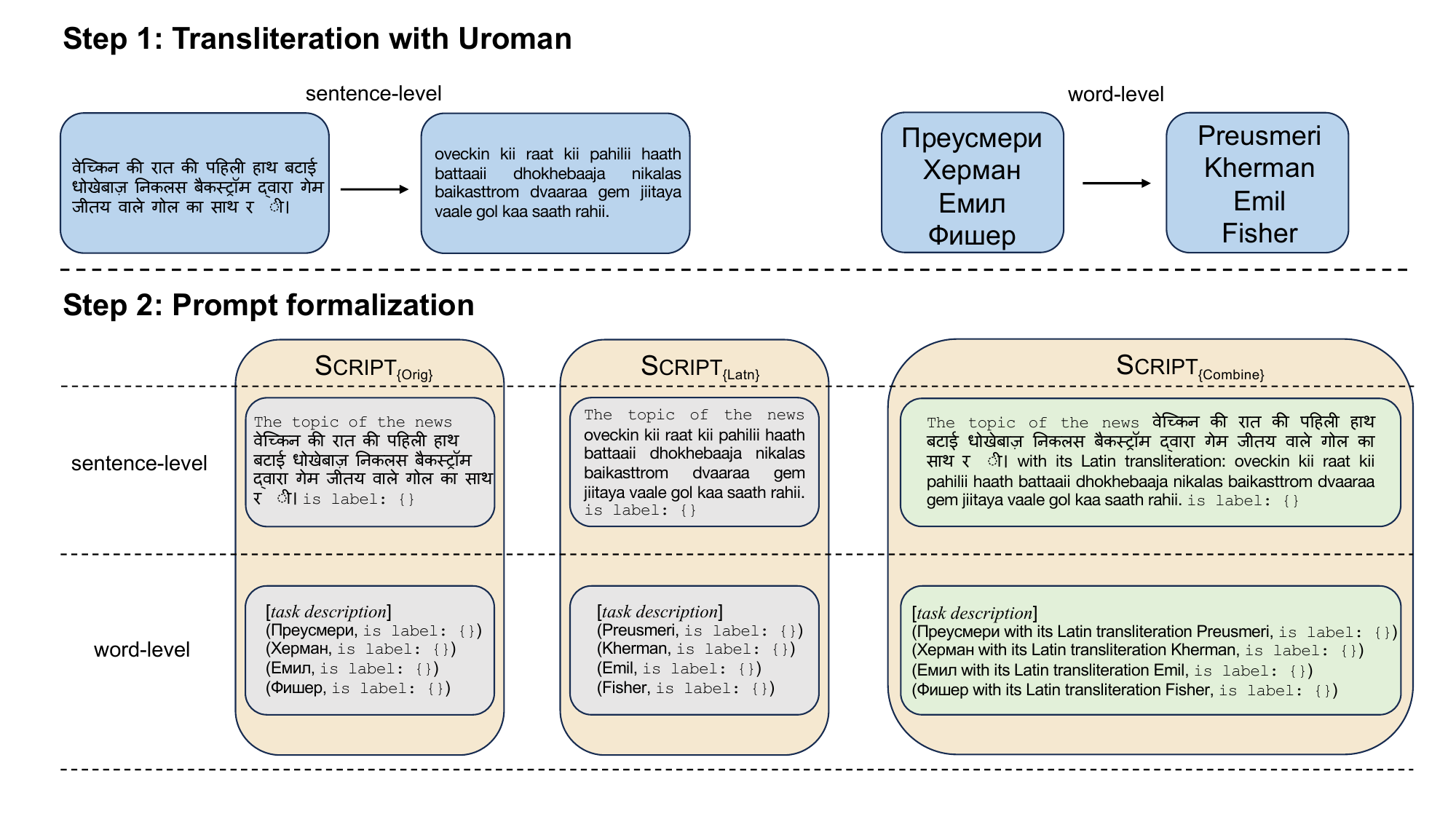}
  \caption{Illustration of our framework. We use \texttt{Uroman} \citep{hermjakob-etal-2018-box} to transliterate non-Latin texts (sentence-level for text classification,
  and word-level for sequential labeling). We propose three prompts: \textbf{\original} (the original text is used), \textbf{\transliteration} (the Latin-script transliteration is used), and \textbf{\combined} (transliteration is used as an augmentation to the original text). }
  \label{fig:framework}
\end{figure*}

Recent studies have demonstrated that leveraging transliteration into a common-script effectively improves the crosslingual transfer performance of encoder-only models on low-resource languages of non-Latin scripts \citep{liu2024translico}. 
This is because a common script facilitates the model to transfer knowledge through increased \textit{lexical overlap} \citep{dhamecha-etal-2021-role, purkayastha-etal-2023-romanization, moosa-etal-2023-transliteration}. 
Inspired by this line of work, a natural research question is to explore whether transliteration is also effective for decoder-only LLMs, especially through their ICL capability which does not require any parameter updates.  

To this end, the paper investigates the above research question and proposes three types of prompt templates where the non-Latin target-language text is represented in (1) its original script (\original), (2) Latin script (\transliteration), or (3) both (\combined).
Given that texts in different scripts convey the same semantics, the knowledge encoded in one script should complement the other. 
A capable model, therefore, should leverage this complementarity: when a word or an entire sentence in the original script is not well understood, the model should refer to its transliteration, and vice versa.
We apply our methods to several LLMs on various tasks and observe that the effectiveness of transliteration varies by task type and model size.
Transliteration is particularly helpful for sequential labeling.
On other tasks, however, transliteration-augmented prompts are less effective, indicating models might have limited capacity to exploit complementary information. 

Our contributions are as follows:
(i) We conduct the first investigation towards the effectiveness of transliteration in ICL for decoder-only LLMs.
(ii) We propose transliteration-augmented prompts that are specifically for low-resource languages in non-Latin scripts; 
(iii) We offer insights on when and how transliteration can enhance ICL performance.


\section{Experimental Settings}

\paragraph{Models.} We experiment with six models: LLaMA2-7B \citep{touvron2023llama}, Mistral-7B-Instruct \citep{jiang2024mixtral}, and the 7B, 3B, 1B, and 560M variants of the BLOOM model \citep{workshop2023bloom}. 
LLaMA2 is a model trained on 28 languages and 5 scripts (Cyrillic, Latin, Hang, Hani and Japanese). 
Mistral is an English-centric model trained on five languages in Latin script, while BLOOM is a multilingual LLM covering a wide range of languages in 11 scripts.\footnote{We check languages covered in each model's training data and consider the dominant script of each language as a script supported by the model.}
We select these models to compare the effectiveness of transliteration-augmented ICL on \textbf{model type} (English-centric vs multilingual models) and \textbf{model size} (different variants of BLOOM).

\begin{table*}[t!]
\setlength{\belowcaptionskip}{-0.3cm}
\centering
\small
\setlength{\tabcolsep}{6mm}
\begin{tabular}{crlrrr}
\toprule
Model & Size & Method & \multicolumn{1}{c}{NER} & \multicolumn{1}{c}{SIB200} & \multicolumn{1}{c}{Taxi1500}
\\ \midrule

\multirow{3}{*}{LLaMA2}
& \multirow{3}{*}{7B}
& \original & \underline{66.8} & \underline{37.2}& \underline{44.8}\\
&&  \transliteration &  \textbf{67.9} & 21.6& \textbf{46.7}\\
&&  \combined &  66.3 &\textbf{48.5} & \textbf{46.7}\\

\midrule

\multirow{3}{*}{Mistral}
& \multirow{3}{*}{7B}
& \original & \underline{69.5} & \textbf{50.6} & \textbf{54.6}\\
&&  \transliteration &  \textbf{71.5} & 33.2 & 51.1\\
&&  \combined & 67.7 &\underline{48.6} & \underline{54.3}\\

\midrule

\multirow{14}{*}{\hfil BLOOM}
& \multirow{3}{*}{7B}
& \original &  65.6 & \textbf{53.5}&\textbf{48.1}\\
&&  \transliteration &  \underline{66.7} &24.3 &45.7\\
&&  \combined &  \textbf{70.0}&\underline{53.2}&\underline{47.4}\\
\cmidrule{2-6}

& \multirow{3}{*}{3B}
& \original & 62.6 & \textbf{48.1}&\textbf{48.0}\\
&&  \transliteration & \underline{63.4} & 29.3&46.5 \\
&&  \combined &  \textbf{68.2} &\underline{39.1} &\underline{47.8}\\
\cmidrule{2-6}

& \multirow{3}{*}{1B}
& \original & 51.6 & \underline{42.4}&\underline{50.3}\\
&&  \transliteration &  \underline{56.5} & 22.0&\textbf{50.4}\\
&&  \combined &  \textbf{64.0} & \textbf{43.8}&\textbf{50.4}\\
\cmidrule{2-6}

& \multirow{3}{*}{560M}
& \original & 52.9 &\textbf{41.5} &\underline{46.1}\\
&&  \transliteration &  \textbf{56.7} &20.4 &45.8\\
&&  \combined & \underline{56.1} &\underline{39.1} &\textbf{46.5} \\
\bottomrule
\end{tabular}
\caption{
Task performance of three prompts (\textbf{\original}, \textbf{\transliteration}, and \textbf{\combined}) for different decoder-only LLMs of various sizes, averaged by languages. Transliteration shows strong effectiveness for NER task but not for other tasks. \textbf{Bold} (\underline{underlined}): best (second-best) result for each model in each task.
}
\label{tab:main_results}
\end{table*}

\paragraph{Methods.} To investigate how transliteration impacts the ICL performance for low-resource languages in non-Latin scripts, we propose three prompt methods: (1) \textbf{\original}, where we feed the model with text in its original script, (2) \textbf{\transliteration}, where we first transliterate the text into Latin script and only feed the transliteration into the model, and (3) \textbf{\combined}, where we combine the text in its original script and its transliteration and feed both together into the model to solve the task. 
The methods are illustrated in Figure \ref{fig:framework}. 
For transliteration, we use \texttt{Uroman} \citep{hermjakob-etal-2018-box}, a tool for universal romanization, which can be applied to any underrepresented scripts with high efficiency. Note that the task description (in English) is the same across all prompt templates. The target-language texts used for few-shot demonstrations are also transliterated in \transliteration and \combined.

\paragraph{Evaluation.}
We consider the following tasks for evaluation:
named entity recognition (\textbf{NER}), a sequence labeling task using WikiANN \citep{pan-etal-2017-cross}; 
\textbf{SIB200} \citep{adelani-etal-2024-sib}, a multilingual classification task covering 205 languages; and 
\textbf{Taxi1500} \citep{ma2024taxi1500}, a
multilingual 6-class text classification dataset contains more than 1,500 languages.
For each task, we only consider a subset of languages that are written in non-Latin scripts (details are shown in \secref{task_info}).
For Taxi1500, we perform a 3-shot prompt and follow the method in \citet{lin2024mala500}, calculating the average of word embeddings in layer
8 of the Glot500 model \citep{imanigooghari-etal-2023-glot500} to retrieve semantically similar ICL samples.
For NER, we perform a 3-shot prompt, since each sentence contains multiple tokens to predict and we find that 3 random demonstrations can usually cover most NER categories.
We perform a 7-shot prompt for SIB200 to ensure the demonstrations cover most classes. Details of selecting the ICL demonstrations are in \secref{prompt_templates}.

\section{Results and Discussion}
\begin{figure*}[ht]
\setlength{\belowcaptionskip}{-0.3cm}
\centering
\subfigure{\includegraphics[width=0.34\textwidth]{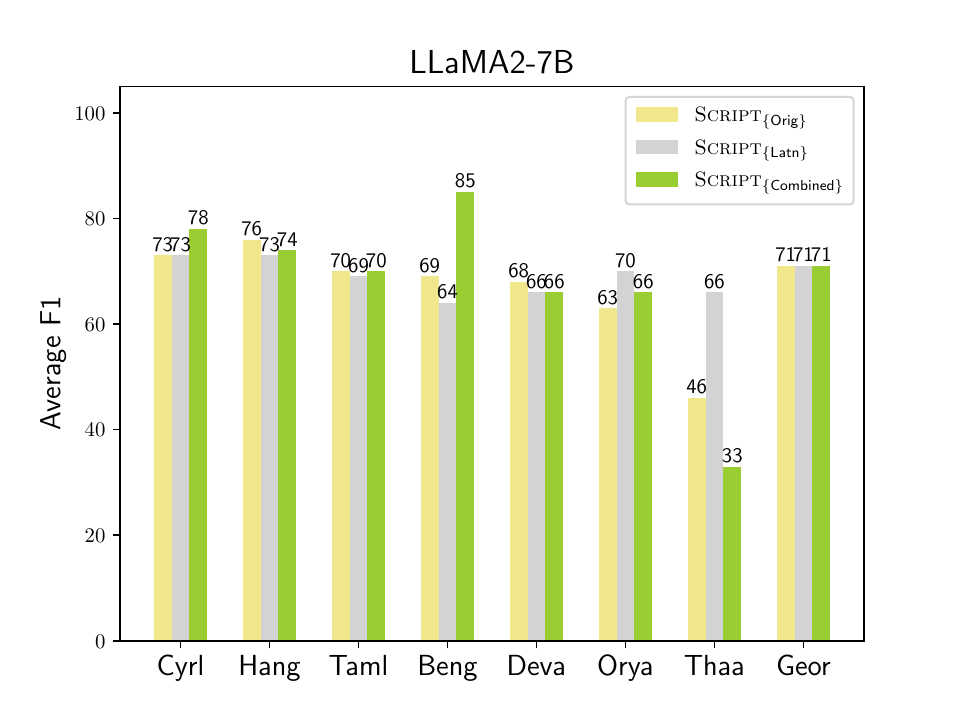}}
\hspace{-0.4cm}
\subfigure{\includegraphics[width=0.34\textwidth]{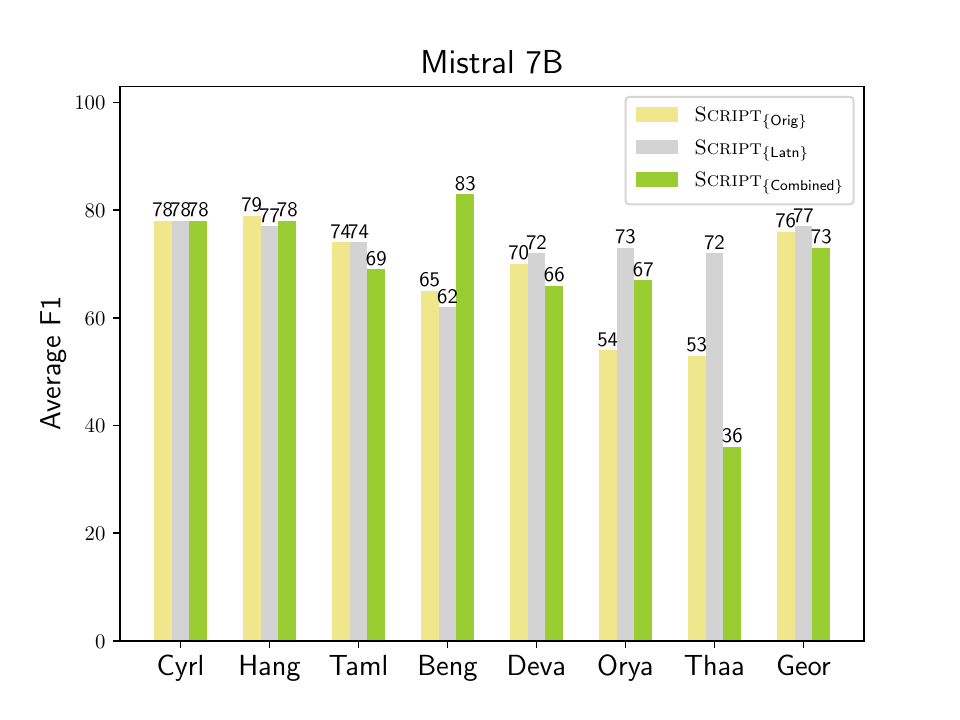}}
\hspace{-0.4cm}
\subfigure{\includegraphics[width=0.34\textwidth]{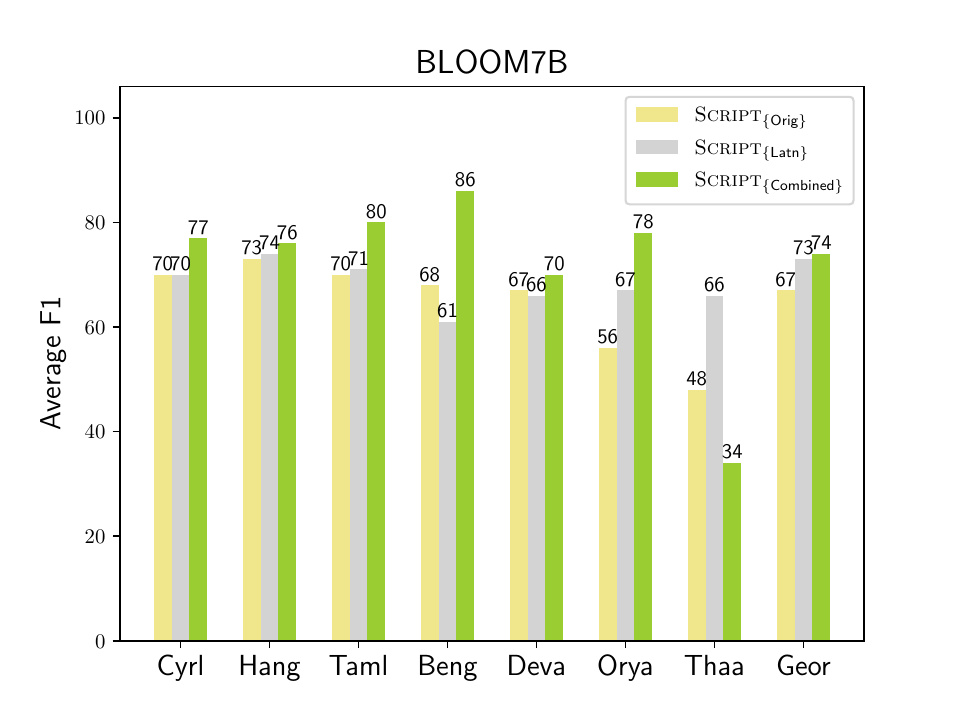}}
\hspace{-0.4cm}
\caption{Performance on NER task averaged by languages of the same script. Transliterations are generally effective in improving the ICL across all models and scripts: \transliteration or \combined outperforms \original.
}
\label{fig:script_avg_sib}
\end{figure*}

We report the average performance across all languages in Table \ref{tab:main_results} (per-language performance is in \secref{full_result}). In addition, we show the performance on NER averaged by script group in Table \ref{fig:script_avg_sib}.

\paragraph{Transliteration benefits sequential labeling.} Across all models, we can observe that either \transliteration or \combined outperforms \original on NER. For instance, \combined increases by 12.4 compared to \original on BLOOM-1B, which is more than 24\% improvement. This demonstrates that models can make better predictions by leveraging the knowledge encoded in the Latin-script transliterations. This can be explained by the fact that NER data contains many (proper) nouns shared across languages. Transliteration enables the model to better exploit such shared vocabularies for inference.


\paragraph{The impact of transliteration on text classification varies across models.} 
\transliteration almost always performs the worst across all models compared with its counterparts, indicating that the transliteration alone is not enough for the model to understand the sentence-level semantics.
Besides, \combined performs suboptimal compared to \original on the English-centric (Mistral) model and models trained on many multilingual data (BLOOM), which suggests these models cannot well leverage complementary information. Instead, such information confuses the models.
However, transliteration can be a good auxiliary input for good Latin-dominant models such as LLaMa (\combined achieves more than 29\% and 4\% on SIB200 and Taxi1500 respectively), as the model can leverage transliteration when it cannot fully understand the text in the original script. 



\paragraph{Model performance varies by different scripts.}
Figure \ref{fig:script_avg_sib} shows the average macro-F1 of ten scripts on the NER task of LLaMA2-7B, Mistral-7B, and BLOOM-7B. For BLOOM-7B, \combined outperforms \original and \transliteration on most scripts except Thaana, a script not seen by BLOOM-7B. Moreover, for scripts covered in the pretraining data (Tamil, Bengali, and Odia), \combined obtains the largest improvement. On the English-centric Mistral-7B, prompts containing transliteration (\transliteration or \combined ) beats \original on 5 out of 8 scripts.
For LLaMA, combining both the original text and transliteration is effective: \combined achieves the best performance on most scripts, indicating a strong ICL capability of exploring commentary information.

\paragraph{Model size plays an important role.} Scaling up the model size usually indicates a stronger capacity from which the ICL can benefit \citep{zhao2023survey}. Indeed, we observe that the performance generally increases for the BLOOM family when the model size scales up for all three prompt types across different tasks except for Taxi1500. We hypothesize this is because Taxi1500 is a relatively easy task and its data builds up on the Bible, which is part of the training data of these LLMs.
In addition, the sentences in Taxi1500 contain many proper nouns whose transliterations the LLMs can easily exploit for making predictions.
Therefore, we also observe good performance for \transliteration (comparable to the other prompts) in Taxi1500, but not in SIB200.

\label{sec:results}

\section{Related Work}
Positive effects of transliterating data into a common script have been demonstrated in various recent works for encoder-only models \citep{dhamecha-etal-2021-role,purkayastha-etal-2023-romanization,moosa-etal-2023-transliteration,liu2024transmi}. Additionally, leveraging transliteration as an auxiliary input at fine-tuning stage improves the cross-script performance \citep{liu2024translico}.
To improve ICL performance for low-resource languages, demonstrations play an important role. One line of approaches replaces the target-language texts with English translations \citep{artetxe-etal-2023-revisiting,Shi2023languagemodel,etxaniz2023multilingual}. Another type of research augments the ICL demonstrations, e.g., by retrieving the most similar English texts to the target-language text \citep{nie-etal-2023-cross, li-etal-2023-crosslingual,wang2023retrievalaugmented}


\section{Conclusion}

This study explores the effectiveness of transliteration in enhancing the ICL performance of decoder-only LLMs, focusing on low-resource languages written in non-Latin scripts.
By proposing three prompt templates -- using original script, Latin script, and a combination of both -- we evaluate their impact across various tasks on several representative LLMs. 
Our findings indicate that transliteration is particularly effective for sequential labeling but its benefits for text classification tasks are less consistent. 
We also observe a mixed effect of transliteration related to the model type and model size.
Our results highlight the potential of transliteration as a possible way to enhance LLMs' performance for low-resource languages.


\section*{Limitations}
There are mainly two limitations in our work.
First, we only consider models with up to 7 billion parameters due to constraints in our computing resources. Second, the evaluation data is limited in terms of the types of tasks.
The major reason is the limited availability of evaluation datasets containing a variety of scripts.
Nevertheless, as a pioneer study in exploring the effectiveness of transliteration for ICL involving low-resource languages in non-Latin scripts, we hope future research can leverage larger models and more datasets to explore this direction.

\bibliography{anthology,custom}

\appendix
\section{Task Data Information}
\seclabel{task_info}

The basic information of each task dataset is shown in
Table \ref{task_evaluation_info}. The number of languages of script groups for each downstream task is shown in Table \ref{evaluation_info_script}.
We introduce the detailed hyperparameters settings for each task in the following.

\begin{figure*}[h!]
    \setlength{\belowcaptionskip}{-0.3cm}
  \centering
  \includegraphics[width=\textwidth]{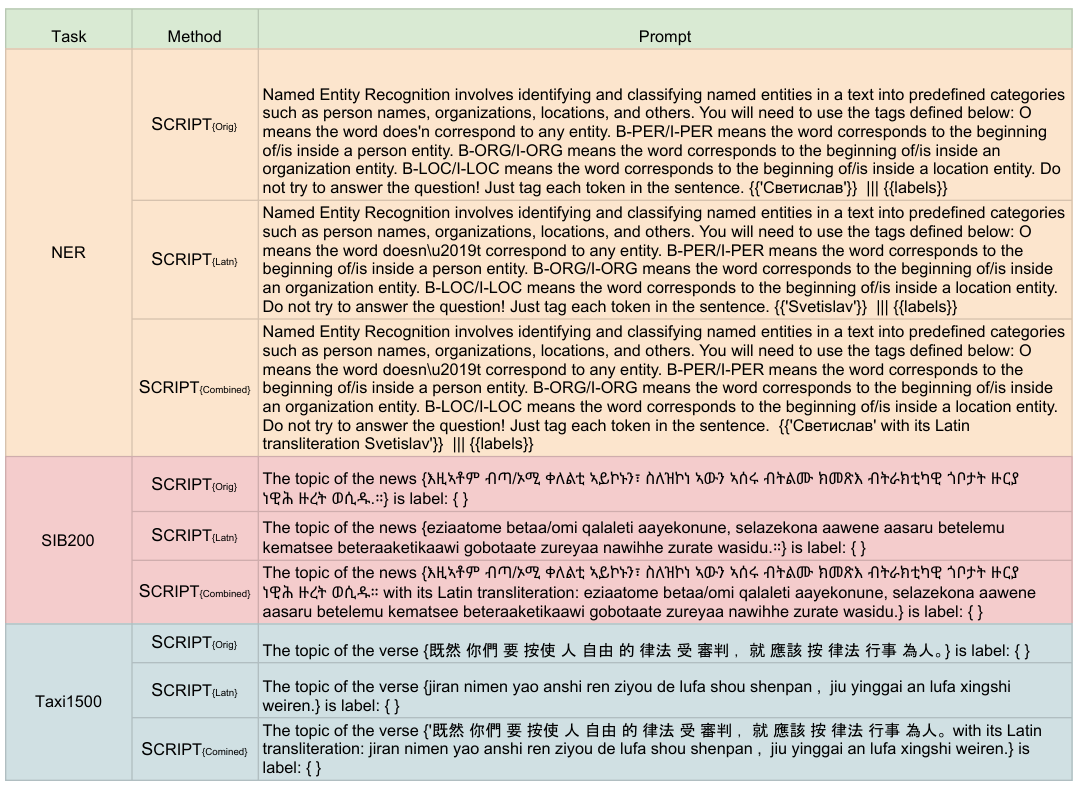}
  \caption{Three types of prompt templates (\original, \transliteration and \combined) that are used for each task. We follow the prompt templates in \citep{lin2024mala500} for the \original, where the target-langauge text is represented in the original script. We use Latin-script transliterations obtained by \texttt{Uroman} \citep{hermjakob-etal-2018-box} for \transliteration. \combined leverages both the original script and its Latin transliteration.}
  \label{fig:prompt_example}
\end{figure*}

For Named Entity Recognition (NER), we employ a 3-shot prompting strategy. Given that each sentence comprises multiple tokens requiring prediction, we have determined that three randomly selected demonstrations typically encompass the majority of NER categories. 
For SIB200, we do a 7-shot prompt. The 7 demonstrations are manually selected to cover the 7 classes of the task. 
For Taxi1500, we use a 3-shot prompt and adhere to the methodology outlined in \citet{lin2024mala500}. 
Specifically, we calculate the average of contextualized word embeddings from layer 8 of the Glot500 model \citep{imanigooghari-etal-2023-glot500} to identify 10 most semantically similar samples, and randomly select 3 samples as the demonstrations.

\begin{table}
  \small
	\centering
	\def\tablesep{0.03cm}
\begin{tabular}{
  @{\hspace{\tablesep}}l@{\hspace{\tablesep}}|
  @{\hspace{\tablesep}}r@{\hspace{\tablesep}}
  @{\hspace{\tablesep}}r@{\hspace{\tablesep}}
  @{\hspace{\tablesep}}r@{\hspace{\tablesep}}
  @{\hspace{\tablesep}}r@{\hspace{\tablesep}}
  @{\hspace{\tablesep}}r@{\hspace{\tablesep}}
  @{\hspace{\tablesep}}r@{\hspace{\tablesep}}
  @{\hspace{\tablesep}}r@{\hspace{\tablesep}}
  @{\hspace{\tablesep}}r@{\hspace{\tablesep}}
  @{\hspace{\tablesep}}r@{\hspace{\tablesep}}
}
 Task&& & & |lan|& & & &&\\
 \midrule
\multirow{2}{*}{\hfil NER} &Cyrl & Arab & Hani & Deva & Geor & Hebr & Beng &other &all\\
&17 & 10 & 5 & 5 & 2 & 2 & 2 & 19 & 62\\
 \midrule
\multirow{2}{*}{\hfil SIB200} &Arab & Deva & Cyrl & Mymr & Beng & Tibt & Hebr & \\
&15 & 9 & 8 & 2 & 2 & 2 & 2 & 22&62\\
 \midrule
\multirow{2}{*}{\hfil Taxi1500} &Cyrl & Arab & Deva & Hani & Mymr & Beng & Orya\\
&24 & 9 & 7 & 3 & 2 & 2 & 2 & 15&64\\
  \end{tabular}
  \caption{The number of languages in each script group for each downstream task.}
  \label{evaluation_info_script}
\end{table}

\begin{table}[t]
  \small
	\centering
	\def\tablesep{0.2cm}
\begin{tabular}{
  @{\hspace{\tablesep}}l@{\hspace{\tablesep}}|
  @{\hspace{\tablesep}}r@{\hspace{\tablesep}}
  @{\hspace{\tablesep}}r@{\hspace{\tablesep}}
  @{\hspace{\tablesep}}r@{\hspace{\tablesep}}
  @{\hspace{\tablesep}}c@{\hspace{\tablesep}}
}
 & |lan| & |rows| & \#class & measure (\%) \\
  \midrule
NER  & 62 & 119 & 7 & F1 score \\
SIB200 & 62 & 1140 & 7 & Accuracy \\
Taxi1500 & 64 & 666 & 6 & Accuracy \\
  \end{tabular}
  \caption{Information of evaluation tasks. |lan|: languages we select as subset to evaluate; \#class: the number of the categories if it is a sequence-level or token-level classification task.}
  \label{task_evaluation_info}
\end{table}


\section{Prompt Templates}
\seclabel{prompt_templates}
We follow the prompt templates in \citep{lin2024mala500} for \original, where the demonstrations and the query are in the original script of the target language. We employ \texttt{Uroman} \citep{hermjakob-etal-2018-box} to transliterate the target-language demonstrations and the target-language query into Latin script. \transliteration only uses the transliteration, while \combined leverage both the original script and its Latin transliteration.

\section{Full Results for All Scripts/Languages}
\seclabel{full_result}

We report the complete results for all tasks and language-scripts in Table \ref{NER_1} and Table \ref{NER_2} (\textbf{NER}), Table \ref{SIB200_1} and  Table \ref{SIB200_2} (\textbf{SIB200}), and Table \ref{Taxi_1} and Table \ref{Taxi_2} (\textbf{Taxi1500}).

\begin{table*}
\centering
\small
\resizebox{\textwidth}{!}{
\begin{tabular}{lrrr|lrrr|lrrr}
\toprule
Language & \original & \transliteration & \combined & Language & \original & \transliteration & \combined & Language & \original & \transliteration & \combined \\
\midrule
ara\_Arab & \textbf{60.4} & \underline{56.0} & 41.1 & ara\_Arab & \underline{63.2} & 59.4 & \textbf{82.3} & ara\_Arab & \underline{69.8} & 63.8 & \textbf{84.4} \\
arz\_Arab & \textbf{62.4} & \underline{58.3} & 40.7 & arz\_Arab & \underline{64.1} & 62.4 & \textbf{77.9} & arz\_Arab & \underline{69.9} & 65.9 & \textbf{81.0} \\
ckb\_Arab & \underline{61.7} & 56.7 & \textbf{73.6} & ckb\_Arab & \underline{58.7} & 57.9 & \textbf{77.7} & ckb\_Arab & \underline{61.6} & 60.4 & \textbf{84.1} \\
fas\_Arab & \textbf{61.9} & \underline{59.7} & 59.4 & fas\_Arab & \underline{62.8} & 61.8 & \textbf{82.6} & fas\_Arab & \underline{66.3} & 63.7 & \textbf{85.1} \\
mzn\_Arab & \textbf{72.8} & \underline{68.6} & 64.2 & mzn\_Arab & 67.2 & \underline{70.0} & \textbf{82.3} & mzn\_Arab & \underline{78.4} & 74.0 & \textbf{87.1} \\
pnb\_Arab & 53.5 & \textbf{58.3} & \underline{58.2} & pnb\_Arab & 57.1 & \underline{57.5} & \textbf{81.0} & pnb\_Arab & \underline{66.4} & 65.8 & \textbf{87.3} \\
pus\_Arab & \textbf{47.2} & \underline{41.2} & 30.4 & pus\_Arab & \textbf{42.4} & \underline{39.5} & 28.3 & pus\_Arab & \textbf{49.3} & \underline{48.7} & 30.2 \\
\bottomrule
\end{tabular}
}
\caption{Macro-F1 score of NER task on BLOOM 560m, BLOOM 1B and BLOOM3B (from left to right).}\label{NER_1}
\end{table*}

\begin{table*}
\centering
\small
\resizebox{\textwidth}{!}{
\begin{tabular}{lrrr|lrrr|lrrr}
\toprule
Language & \original & \transliteration & \combined & Language & \original & \transliteration & \combined & Language & \original & \transliteration & \combined \\
\midrule
ara\_Arab & \underline{70.6} & 65.7 & \textbf{86.9} & ara\_Arab & \underline{69.8} & 65.5 & \textbf{86.0} & ara\_Arab & \underline{74.0} & 69.2 & \textbf{85.9} \\
arz\_Arab & \underline{69.8} & 67.7 & \textbf{81.3} & arz\_Arab & \textbf{73.6} & 67.0 & \underline{70.1} & arz\_Arab & \underline{78.0} & 73.1 & \textbf{78.7} \\
ckb\_Arab & \underline{64.0} & 63.0 & \textbf{84.7} & ckb\_Arab & \underline{65.5} & 64.2 & \textbf{82.0} & ckb\_Arab & \underline{64.3} & 62.2 & \textbf{83.7} \\
fas\_Arab & \underline{70.5} & 67.3 & \textbf{87.2} & fas\_Arab & \underline{73.1} & 68.2 & \textbf{85.9} & fas\_Arab & \underline{70.8} & 67.8 & \textbf{84.5} \\
mzn\_Arab & \underline{80.4} & 79.4 & \textbf{87.7} & mzn\_Arab & \underline{79.7} & 77.9 & \textbf{82.3} & mzn\_Arab & \underline{81.9} & 81.3 & \textbf{83.9} \\
pnb\_Arab & 68.5 & \underline{68.7} & \textbf{86.9} & pnb\_Arab & \underline{72.5} & 71.1 & \textbf{90.2} & pnb\_Arab & \underline{74.0} & 71.7 & \textbf{90.9} \\
pus\_Arab & \underline{51.1} & \textbf{52.9} & 31.7 & pus\_Arab & \underline{53.0} & \textbf{58.6} & 33.9 & pus\_Arab & \underline{55.5} & \textbf{57.1} & 29.4 \\
\bottomrule
\end{tabular}
}
\caption{Macro-F1 score of NER task on NER task on BLOOM 7B, LLaMA2-7B and  Mixtral 7B (from left to right)}\label{NER_2}
\end{table*}

\begin{table*}
\centering
\small
\resizebox{\textwidth}{!}{
\begin{tabular}{lrrr|lrrr|lrrr}
\toprule
Language & \original & \transliteration & \combined & Language & \original & \transliteration & \combined & Language & \original & \transliteration & \combined \\
\midrule
ace\_Arab & 18.1 & \underline{18.6} & \textbf{20.6} & ace\_Arab & 16.7 & \underline{16.7} & \textbf{21.6} & ace\_Arab & \textbf{27.5} & 19.6 & \underline{20.6} \\
acm\_Arab & \underline{63.7} & 16.7 & \textbf{66.7} & acm\_Arab & \underline{67.6} & 18.1 & \textbf{70.1} & acm\_Arab & \textbf{77.9} & 19.1 & \underline{66.7} \\
acq\_Arab & \underline{63.7} & 16.7 & \textbf{64.7} & acq\_Arab & \underline{69.1} & 17.2 & \textbf{72.1} & acq\_Arab & \textbf{77.9} & 21.1 & \underline{64.7} \\
aeb\_Arab & \textbf{63.2} & 17.6 & \underline{56.9} & aeb\_Arab & \underline{62.7} & 17.2 & \textbf{67.2} & aeb\_Arab & \textbf{74.0} & 18.6 & \underline{56.9} \\
ajp\_Arab & \textbf{68.6} & 18.1 & \underline{67.2} & ajp\_Arab & \underline{71.1} & 18.1 & \textbf{74.5} & ajp\_Arab & \textbf{75.0} & 25.5 & \underline{67.2} \\
apc\_Arab & \underline{70.1} & 19.1 & \textbf{71.1} & apc\_Arab & \underline{74.5} & 17.2 & \textbf{75.0} & apc\_Arab & \textbf{77.9} & 26.5 & \underline{71.1} \\
ars\_Arab & \textbf{65.7} & 15.7 & 65.7 & ars\_Arab & \underline{67.6} & 16.7 & \textbf{72.1} & ars\_Arab & \textbf{76.5} & 19.6 & \underline{65.7} \\
\bottomrule
\end{tabular}
}
\caption{Accuracy of SIB200 task on BLOOM 560m, BLOOM 1B and BLOOM3B (from left to right).}\label{SIB200_1}
\end{table*}

\begin{table*}
\centering
\small
\resizebox{\textwidth}{!}{
\begin{tabular}{lrrr|lrrr|lrrr}
\toprule
Language & \original & \transliteration & \combined & Language & \original & \transliteration & \combined & Language & \original & \transliteration & \combined \\
\midrule
ace\_Arab & \underline{22.1} & 17.6 & \textbf{24.5} & ace\_Arab & \underline{17.6} & 11.8 & \textbf{19.6} & ace\_Arab & \underline{29.4} & 16.7 & \textbf{29.9} \\
acm\_Arab & \underline{79.9} & 10.3 & \textbf{81.4} & acm\_Arab & \underline{63.7} & 22.5 & \textbf{70.1} & acm\_Arab & \textbf{77.0} & 22.5 & \underline{73.0} \\
acq\_Arab & \underline{78.9} & 12.7 & \textbf{81.9} & acq\_Arab & \underline{62.3} & 15.2 & \textbf{67.2} & acq\_Arab & \textbf{77.0} & 21.1 & \underline{74.0} \\
aeb\_Arab & \textbf{76.5} & 12.3 & \underline{74.5} & aeb\_Arab & \underline{58.8} & 17.6 & \textbf{65.7} & aeb\_Arab & \textbf{72.1} & 20.1 & \underline{69.6} \\
ajp\_Arab & \textbf{82.4} & 19.1 & \underline{76.5} & ajp\_Arab & \underline{60.3} & 17.2 & \textbf{65.7} & ajp\_Arab & \textbf{72.5} & 26.0 & \underline{70.1} \\
apc\_Arab & \underline{79.9} & 17.6 & \textbf{81.9} & apc\_Arab & \underline{56.9} & 16.2 & \textbf{65.2} & apc\_Arab & \textbf{75.5} & 24.0 & \underline{73.5} \\
ars\_Arab & \underline{78.4} & 10.8 & \textbf{78.9} & ars\_Arab & \underline{61.3} & 15.2 & \textbf{71.1} & ars\_Arab & \textbf{77.5} & 20.1 & \underline{75.0} \\
\bottomrule
\end{tabular}
}
\caption{Accuracy of SIB200 task on BLOOM 7B, LLaMA2-7B and  Mixtral 7B (from left to right)}\label{SIB200_2}
\end{table*}

\begin{table*}
\centering
\small
\resizebox{\textwidth}{!}{
\begin{tabular}{lrrr|lrrr|lrrr}
\toprule
Language & \original & \transliteration & \combined & Language & \original & \transliteration & \combined & Language & \original & \transliteration & \combined \\
\midrule
arb\_Arab & 45.9 & \underline{45.9} & \textbf{46.8} & arb\_Arab & 53.2 & \textbf{54.1} & \underline{53.2} & arb\_Arab & \underline{49.5} & 46.8 & \textbf{55.0} \\
ary\_Arab & 34.2 & \textbf{41.4} & \underline{36.9} & ary\_Arab & 43.2 & \underline{45.0} & \textbf{45.9} & ary\_Arab & \underline{36.0} & 35.1 & \textbf{44.1} \\
arz\_Arab & 35.1 & \underline{35.1} & \textbf{36.9} & arz\_Arab & 44.1 & \textbf{45.9} & 45.9 & arz\_Arab & \underline{40.5} & 36.9 & \textbf{41.4} \\
azb\_Arab & \textbf{43.2} & \underline{42.3} & 39.6 & azb\_Arab & 46.8 & \textbf{50.5} & \underline{48.6} & azb\_Arab & 41.4 & \underline{41.4} & \textbf{43.2} \\
ckb\_Arab & 45.0 & \underline{46.8} & \textbf{47.7} & ckb\_Arab & 46.8 & \textbf{48.6} & 48.6 & ckb\_Arab & 43.2 & \textbf{44.1} & \underline{43.2} \\
fas\_Arab & \textbf{53.2} & 49.5 & 53.2 & fas\_Arab & 53.2 & \textbf{55.0} & 55.0 & fas\_Arab & \textbf{49.5} & \underline{49.5} & 49.5 \\
pes\_Arab & \underline{53.6} & 46.4 & \textbf{55.5} & pes\_Arab & 50.0 & \underline{51.8} & \textbf{55.5} & pes\_Arab & \underline{49.1} & 48.2 & \textbf{50.0} \\
\bottomrule
\end{tabular}
}
\caption{Accuracy of Taxi1500 task on BLOOM 560m, BLOOM 1B and BLOOM3B (from left to right).}\label{Taxi_1}
\end{table*}

\begin{table*}
\centering
\small
\resizebox{\textwidth}{!}{
\begin{tabular}{lrrr|lrrr|lrrr}
\toprule
Language & \original & \transliteration & \combined & Language & \original & \transliteration & \combined & Language & \original & \transliteration & \combined \\
\midrule
arb\_Arab & \underline{49.5} & 45.9 & \textbf{51.4} & arb\_Arab & 43.2 & \textbf{45.9} & 45.9 & arb\_Arab & \textbf{62.2} & 48.6 & \underline{61.3} \\
ary\_Arab & \textbf{38.7} & 30.6 & 38.7 & ary\_Arab & \underline{32.4} & \textbf{35.1} & 29.7 & ary\_Arab & \textbf{55.9} & 38.7 & \underline{50.5} \\
arz\_Arab & \textbf{45.0} & 35.1 & \underline{41.4} & arz\_Arab & 31.5 & \textbf{39.6} & \underline{34.2} & arz\_Arab & \underline{54.1} & 47.7 & \textbf{58.6} \\
azb\_Arab & \underline{47.7} & 43.2 & \textbf{48.6} & azb\_Arab & 39.6 & \textbf{42.3} & 42.3 & azb\_Arab & \underline{51.4} & 42.3 & \textbf{55.9} \\
ckb\_Arab & \underline{45.0} & \textbf{47.7} & 42.3 & ckb\_Arab & \underline{44.1} & \textbf{45.0} & 42.3 & ckb\_Arab & \textbf{47.7} & 44.1 & \underline{46.8} \\
fas\_Arab & \textbf{57.7} & \underline{51.4} & 49.5 & fas\_Arab & 49.5 & \underline{50.5} & \textbf{53.2} & fas\_Arab & \textbf{66.7} & 46.8 & \underline{63.1} \\
pes\_Arab & \textbf{59.1} & 51.8 & \underline{56.4} & pes\_Arab & \underline{50.9} & 49.1 & \textbf{56.4} & pes\_Arab & \textbf{64.5} & 49.1 & \underline{62.7} \\
\bottomrule
\end{tabular}
}
\caption{Accuracy of Taxi1500 task on BLOOM 7B, LLaMA2-7B and  Mixtral 7B (from left to right).}\label{Taxi_2}
\end{table*}

\end{document}